\pdfoutput=1

\documentclass[11pt]{article}

\usepackage[]{emnlp2021}

\usepackage{times}
\usepackage{latexsym}

\usepackage{url}
\usepackage{subfigure}
\usepackage{graphicx}
\usepackage[normalem]{ulem}
\usepackage{paralist}
\usepackage{multirow}

\usepackage[T1]{fontenc}

\usepackage[utf8]{inputenc}

\usepackage{microtype}

\title{Who Responded to Whom: The Joint Effects of Latent Topics and Discourse in Conversation Structure}

\author{First Author \\
  Affiliation / Address line 1 \\
  Affiliation / Address line 2 \\
  Affiliation / Address line 3 \\
  \texttt{email@domain} \\\And
  Second Author \\
  Affiliation / Address line 1 \\
  Affiliation / Address line 2 \\
  Affiliation / Address line 3 \\
  \texttt{email@domain} \\}

\author { Lu Ji$^1$, 
Jing Li$^{2}$,
Zhongyu Wei$^3$, 
Qi Zhang$^1$, 
Xuanjing Huang$^1$\\
$^1$School of Computer Science, Fudan University, Shanghai, China 
\\ $^2$Department of Computing, The Hong Kong Polytechnic University, Hong Kong, China  
\\ $^3$School of Data Science, Fudan University, China\\
{\texttt  \{17210240034,zywei,qz,xjhuang\}@fudan.edu.cn } \\
{\texttt jing-amelia.li@polyu.edu.hk}
}

\begin{document}
\maketitle
\begin{abstract}
Numerous online conversations are produced on a daily basis, resulting in a pressing need to conversation understanding.
As a basis to structure a discussion, we identify the responding relations in the conversation discourse, which link response utterances to their initiations.
To figure out who responded to whom, here we explore how the consistency of topic contents and dependency of discourse roles indicate such interactions, whereas most prior work ignore the effects of latent factors underlying word occurrences.
We propose a model to learn latent topics and discourse in word distributions, and predict pairwise initiation-response links via exploiting topic consistency and discourse dependency.
Experimental results on both English and Chinese conversations show that our model significantly outperforms the previous state of the arts, such as $79$ vs. $73$ MRR on Chinese customer service dialogues.
We further probe into our outputs and shed light on how topics and discourse indicate conversational user interactions.
\end{abstract}
\section{Introduction}

The growing popularity of online platforms have resulted in the revolution of interpersonal communications.
Individuals now engage in diverse forms of online conversations to exchange viewpoints and share ideas.
It allows users to access an abundance of fresh materials, whereas the explosive growth of online texts --- essentially conversational and usually in multiple threads~\cite{wang:2010} --- has also hindered human capability to find the information needed.
There consequently presents a pressing need to develop conversation understanding methods to digest massive texts and complex interactions therein. 
To that end, it is crucial to capture the interactions of who responded to whom --- the base to build and understand the conversation structure, as pointed out in many previous studies~\cite{wang:2010}.
By reflecting how participants interact with each other, such structure has shown useful to predict users' online social activities~\cite{DBLP:conf/emnlp/ZengLWW19}, summarize key discussion topics~\cite{DBLP:journals/coling/LiSWW18}, measure argument persuasiveness~\cite{DBLP:conf/coling/JiWHLZH18}, and so forth.

\begin{figure}[t]
\footnotesize 
\begin{tabular}{p{0.455\textwidth}}
\hline
\multicolumn{1}{|p{0.455 \textwidth}|}{[$C_1$] I am aware that  you can \textbf{thank} them in \textbf{private argument} but what does \textcolor{blue}{\textit{that}} matter?} \\
\multicolumn{1}{|p{0.455 \textwidth}|}{[$C_2$] The most important part of my argument is that it \textbf{hurts} literally nobody.} \\
\multicolumn{1}{|p{0.455 \textwidth}|}{[$C_3$] All they are doing is trying to be \textbf{polite}. } \\
\multicolumn{1}{|p{0.455 \textwidth}|}{[$C_4$] Some people gild comments anonymously and do not respond to the \textbf{private messages}, so the gildee never knows who gave them gold. } \\
\multicolumn{1}{|p{0.455 \textwidth}|}{[$C_5$] Note: for the purposes of my argument, assume I am talking about comments edited in such a way as to say \textbf{thanks} for the gold!} \\
\hline
\hline
\multicolumn{1}{|p{0.455 \textwidth}|}{[$R$] We are all aware that you can do \textcolor{blue}{\textit{that}}, but sometimes people like to \textbf{express gratitude publicly}.}\\
\hline
\end{tabular}
\vspace{-0.5em}
\caption{\label{fig11} A Reddit conversation snippet. $C_1$ and $R$ is an initiation-response pair while $C_2$ to $C_5$ are the other four candidates.
\textbf{Topic words} reflecting the discussion points ``public gratitude expression'' are in bold. The blue and italic ``\textcolor{blue}{\textit{that}}'' occurring in both $C_1$ and $R$ imply $R$'s possible intention to answer $C_1$'s question.
}
\vspace{-2em}
\end{figure}

To date, despite of the extensive efforts on user interaction modeling, many of them employ user-annotated in-reply-to signals, such as \textit{@-mention} on Twitter~\cite{DBLP:journals/coling/LiSWW18,DBLP:conf/emnlp/ZengLWW19}. 
Nonetheless, such labels are usually unavailable or unreliable~\cite{du2017discovering,he:2019}, especially for online conversations in informal styles.
Other studies assume utterances only respond to their chronological neighbors~\cite{DBLP:conf/www/JiaoL0M18,zhao2018unsupervised}, largely ignoring the long-distance interactions prominent in online conversations~\cite{wang:2010}.
All these concerns lay down our objective to investigate who responded to whom in conversation contexts.

Following previous practice~\cite{Schegloff:2007}, we define our task to predict pairwise initiation utterances and their responses in an online conversation (henceforth \textbf{initiation-response pairs}), where an initiation sets up an expectation earlier and its response later react to it in process of a discussion.
To illustrate our task, Figure \ref{fig11} shows an example response $R$ and the other five utterances $C_1$ to $C_5$ from $R$'s previous post in a Reddit conversation.
Our goal is to identify which utterance from $C_1$ to $C_5$ is $R$'s initiation.
As can be seen, $R$ is most likely to respond to $C_1$ for two possible reasons: 
First, they both focus on the topic of \textit{public gratitude expression} (as topic words ``\textit{thank}'', ``\textit{public}'', ``\textit{gratitude}'' are mentioned); 
Second, $C_1$ raises a question (signaled by ``\textit{what}'' and the question mark ``\textit{?}'') that can be well answered by $R$ (via echoing the pronoun ``\textit{that}'').

Here, we examine two latent factors that implicitly link an initiation and its response --- the consistency of the topics they center around (henceforth \textbf{topic consistency}) and the dependency of their discourse roles (henceforth \textbf{discourse dependency}).\footnote{Discourse roles refer to utterance-level dialogue acts, such as asking a question and making an argument~\cite{ritter2010unsupervised}. }
Our intuition is that responses tend to follow the points pushed forward in their initiations (such as \textit{public gratitude expression} in Figure \ref{fig11}) and their discourse roles are likely to exhibit dependency in interactions, such as an answer responding to an initiated question (like $R$ answering $C_1$ in Figure \ref{fig11}) and an argument followed by another argument in a back-and-forth debate.
To the best of our knowledge, \emph{we are the first to analyze the effects of topics and discourse in conversational responding behavior}, while previous work predict initiation-response pairs without modeling such latent factors embedded in the relations~\cite{du2017discovering}.

To learn topics and discourse, we separate two word distributions for representing each of them. The latent variables are inferred with a neural architecture in an unsupervised manner~\cite{zeng2019you}, which enables topic and discourse inference without either manually annotated data~\cite{zhao2017learning} or expertise involvements to customize model inference~\cite{DBLP:journals/coling/LiSWW18}. 
Afterwards, two neural modules are employed, one to capture topic consistency and the other discourse dependency, both aim to explore the implicit links of a response and a candidate initiation.
The learned representations are hence coupled to predict how likely the two utterances form an initiation-response pair. 

In an empirical study, we conduct extensive experiments on two conversation datasets, one contains English argumentative discussions on Reddit (from the ChangeMyView subreddit), and the other Chinese customer service dialogues from e-commerce platform \textit{Wangwang}. 
Both of them will be released upon publication as part of our work.
The experimental results show that our model significantly outperforms state-of-the-art methods from previous work.
For example, we achieve $79.02$ MRR on the Wangwang dialogues compared with $72.69$ produced by \citet{he:2019}.
In extensive analyses on latent topics and discourse, we find that meaningful representations can be learned by our model and both topics and discourse may contribute to indicate initiation-response pairs.
Lastly, we show that our learned representation to indicate initiation-response relations can benefit to identify persuasive arguments in social media debates.

\section{Study Design}

\subsection{Task Formulation}

We define initiation-response pairs following \citet{Schegloff:2007} and refer both initiations and responses to conversation utterances from different participants.
In a discussion flow, responses appear and react to the points raised earlier in their initiations and hence hold responding relations with them. 
Here, we formulate our task into pairwise ranking following previous settings \cite{wang:2010} and will experiment on two types of initiation-response pairs: quotation-reply pairs in forum discussions and question-answer pairs in customer service dialogues.
Specifically, given a response utterance $r$, we rank a set of candidate utterances with one positive initiation $q^+$ and $u$ negative ones $q_1^-\sim q_u^-$.
In practice, we measure a matching score $S(q,r)$ to indicate the likelihood of $q$ as $r$'s initiation and the one with the highest score will be considered as $r$'s predicted initiation.

\subsection{Data Collection and Analysis}\label{ssec:data}

Two conversation datasets are collected for experiments --- a \textit{CMV} dataset in English from the \textit{CMV} subreddit and the other in Chinese named as \textit{CS} from \textit{Wangwang} customer service platform. \textit{CS} dialogues are synchronous while \textit{CMV} asynchronous.

\paragraph{Data Collection.} \textit{CMV} gathers social media arguments, whose raw data is released by \citet{Tan:16}. For each discussion, we examine the context of an opinion holder (\textit{OH})'s post and a challenger's comment to explore the quotation-reply relations therein.
In challenger's comments, we form a quotation and the utterance right after it to be an initiation-response pair. The rest utterances in the quoted post (from \textit{OH}) are used as the negative instances, and the samples are randomly selected with a cap at $4$ to avoid unbalanced labels. 


\textit{CS} is released with \citet{he:2019}. In a dialogue thread, customers may raise multiple questions sequentially and the seller's answers may appear in the following turns. Our goal is to pair a question from the customer's utterances and an answer from the seller's. The newest $4$ consecutive utterances from the customer (skipping the positive initiation) before a seller's response serve as the negative instances. Here the candidate number is also capped at $4$ for comparable results with \textit{CMV}.

\begin{table}[t!]\footnotesize
\begin{center}
\begin{tabular}{|l|c|c| }
\hline
 & CMV Dataset & CS Dataset \\ \hline
\# of utt. per conv & 21.2$\pm$15.6   & 9.6$\pm$2.8 \\
\# of words per conv  & 403.1$\pm$292.5 & 130.8$\pm$73.1\\
\# of convs & 7,937 & 4,277 \\
\hline
\# of words per $r$  & 19.7$\pm$6.0 &  15.0$\pm$20.8 \\
\# of words per $q^+$ &20.6$\pm$6.2 & 6.5$\pm$4.3\\
\# of words per $q^-$ &16.5$\pm$5.0 & 11.2$\pm$18.7\\ \hline
max \# of pairs   &14  & 7\\ 
avg. \# of pairs  &1.1$\pm$0.3 & 1.7$\pm$1.1\\\hline
\end{tabular}
\end{center}
\vspace{-1em}
\caption{\label{tab:tb1} Data statistics. Means and standard deviations appear before and after $\pm$. $r$ refers to response, while $q^+$ and $q^-$ for positive and negative initiation. \# of pairs represents the number of initiation-response pairs per conversation.}
\end{table}

\begin{figure}[t!]
\centering
  \includegraphics[width=0.75\columnwidth,,height=2.7cm ]{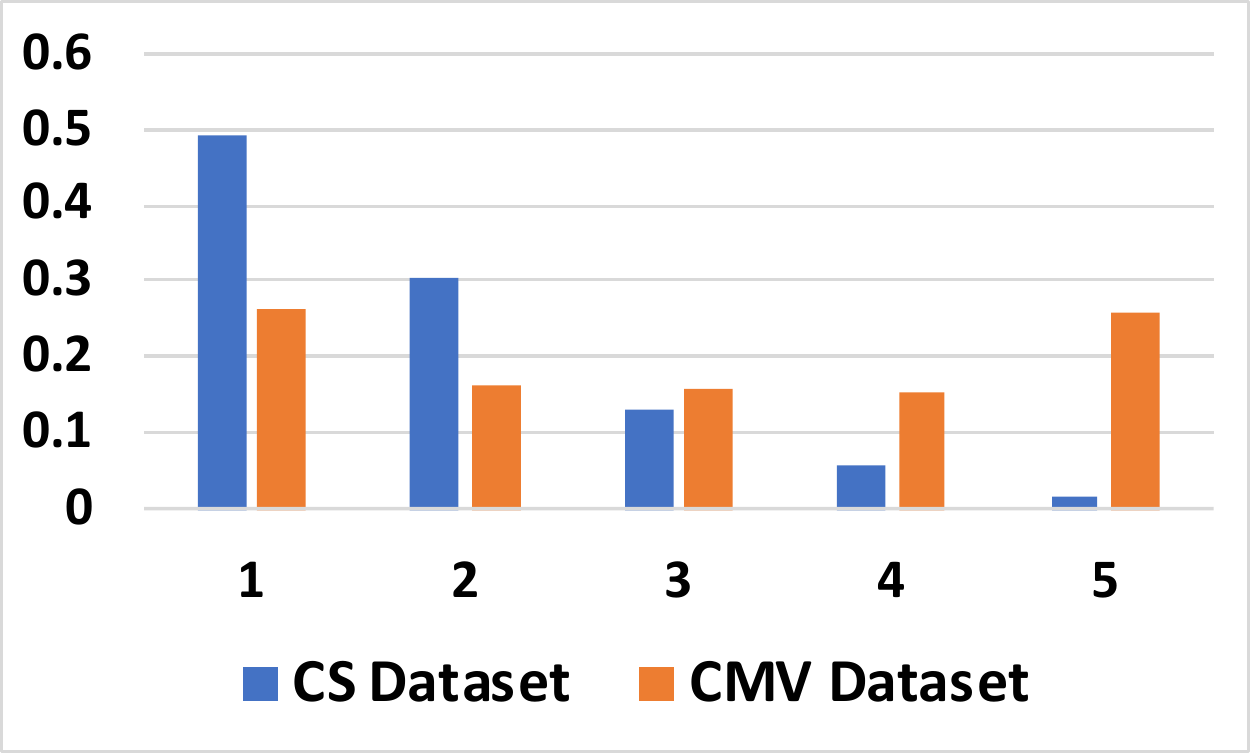}
     \vspace{-0.5em}
    \caption{\label{fig_sta}The distribution over relative positions of initiations and responses. 
    X-axis: initiations' utterance order counted from responses (only considering customer's or \textit{OH}'s turns). Y-axis: proportions.}
\vspace{-1em}
\end{figure}

\paragraph{Data Analysis.} Table \ref{tab:tb1} shows the data statistics, where the two datasets exhibit different characteristics. 
We further analyze the relative positions of initiations and responses and show the distribution of their intermediate utterance number in Figure \ref{fig_sta}. 
As can be seen, large proportion of responses do not interact with the closest utterance, though \textit{CS} sellers do respond more to newer questions, probably because of recency effects in in synchronous dialogues --- people's attention tends to be drawn by new information.
However, in asynchronous forum discussions, \textit{CMV} challengers are more likely to quote the opening points in \textit{OH}'s post.
Another possible reason is that most key arguments are located at the beginning of a post.

\section{Learning Topics and Discourse Effects for Initiation-Response Prediction}
\begin{figure*}[t!]
\subfigure{
\includegraphics[width=0.8\columnwidth] {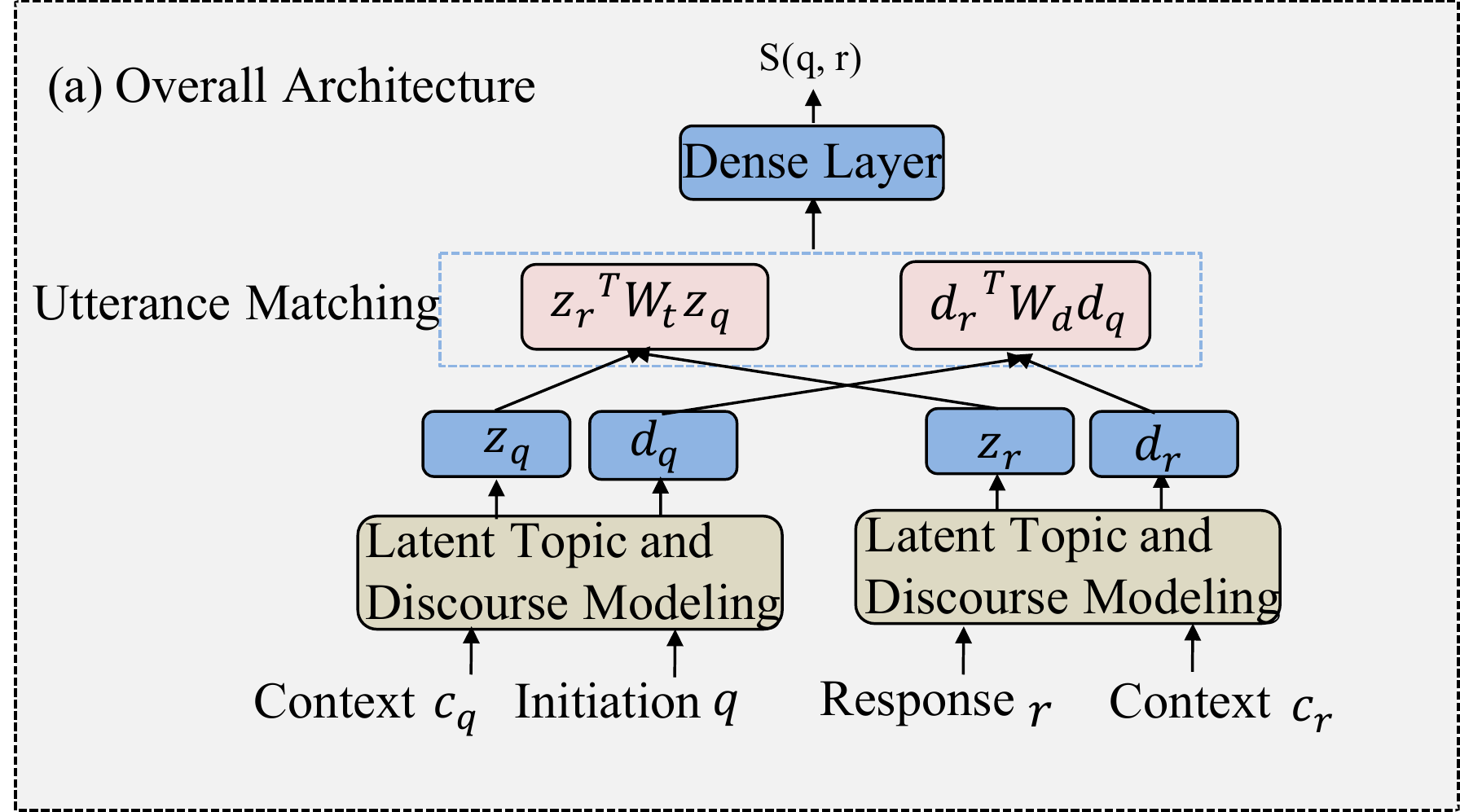}
}
\subfigure{
\includegraphics[width=1.1\columnwidth] {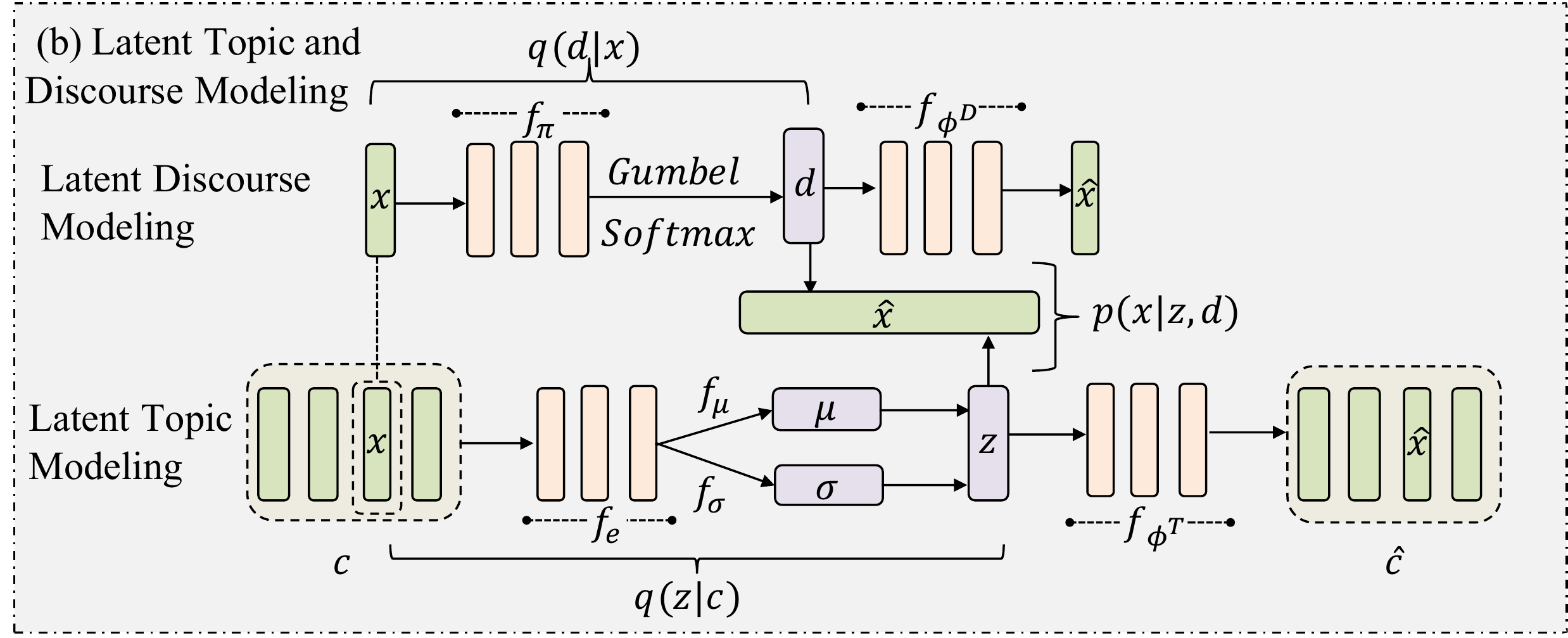}
}
\caption{ \label{figure2} (a) Our model architecture to predict initiation-response pairs. We first learn latent topics and discourse factors for both response $r$ and the candidate initiation $q$ in award of their contexts $c_r$ and $c_q$ and show the detailed learning process in (b) ($x$ denotes $q$ or $r$.) Then, utterance matching is conducted to measure topic consistency and discourse dependency. Lastly, we predict $S(q,r)$ --- the likelihood of $r$ responding to $q$.}
\end{figure*}

The overall architecture of our model is shown in Figure~\ref{figure2} (a). 
It takes an initiation candidate $q$, a response $r$, and their corresponding contexts $c_q$ and $c_r$ as inputs. 
The outputs are matching scores indicating how likely $r$ responds to $q$. 
The model mainly consists of two modules --- one jointly models latent topics and discourse (to be described in Section \ref{ssec:model:topic-disc}) in both $q$ and $r$ (both denoted by utterance $x$ and its context by $c$) and the other predict initiation-response relations coupling topic consistency and discourse dependency (to be covered in Section \ref{ssec:model:pair-pred}). 
The learning objectives will be presented at last in Section \ref{ssec:model:objective}.

\subsection{Latent Topics and Discourse Modeling}\label{ssec:model:topic-disc}

Inspired by previous efforts in neural topic models~\cite{miao:2017}, we adopt variational autoencoder (VAE)~\cite{Kingma:13} to learn latent topics and discourse. 
It allows their associated word distributions to be learned in neural architecture and end-to-end training with other components in a deep learning framework. 
The corresponding networks are illustrated in Figure \ref{figure2} (b).
In below, we first describe how we model the topics, followed by the process to learn discourse. 

\paragraph{Latent Topics.}
We first assume there are $K$ latent topics in the corpus, each represented by a word distribution $\Phi_k^{T} (k = 1,2,...,K)$ over the vocabulary $V$. 
The latent topics of each utterance is defined as $z$ and generated from the topic composition of its  context $c$. 
Here we learn utterance-level topics in its conversation context assuming that utterances in a discussion excerpt tend to focus on similar topics. 
It allows the modeling of rich patterns of word statistics for topic inference.

The following process presents how to generate an utterance $x$ in context of $c$. 
Here, we adopt the bag-of-words assumption of most latent topic models~\cite{blei2002latent,miao:2017} and generate $x$ in its bag of words (BoW) form $x^{BoW}$.

\begin{compactitem}
\item Draw the latent topic $z \sim N (\mu, \sigma^2)$
\item $c$'s topic mixture $\theta = softmax(f_\theta(z))$
\item For the $n$-th word in $x$:
\begin{compactitem}
 \item $\beta_n = softmax(f_{\Phi^T}(\theta))$
 \item Draw the word $w_n \sim Multi(\beta_n)$ 
\end{compactitem}
\end{compactitem}
\noindent where $f_{*}(\cdot)$ is a neural perceptron. The weight matrix of $f_{\Phi^T} (\cdot)$ (after the softmax normalization) is viewed as the topic-word distributions $\Phi^T$. 

The prior parameters $\mu$ and $\sigma$ are estimated from conversation $c$'s bag of words $c^{BoW}$:

\begin{equation}\small
\setlength{\abovedisplayskip}{-1pt}
\mu=f_{\mu}(f_e(c^{BoW})), \,  \log\sigma= f_{\sigma}(f_e(c^{BoW}))
\end{equation}
\noindent $f_{\mu}$, $f_e$ and $f_{\sigma}$ are neural perceptron defined above.

As can be seen, the entire topic modeling process follows a VAE fashion --- for each utterance $x$, we first encode its latent topic $z$ from the conversation context $c$ (in BoW form $c^{BoW}$) and then reconstruct its BoW ($x^{BoW}$) via decoding. 

\paragraph{Latent Discourse.}

Similar to latent topics, we represent latent discourse with word distributions $\Phi_d^D$ ($d=1,2,...,D$) and   
$D$ denotes the number of discourse roles observed from the corpus.

Following \citet{ritter2010unsupervised}, we assume each utterance $x$ reflects only one discourse role $d$ (to signal its dialogue act).
It is hence represented by a $D$-dimensional one-hot vector over the discourse inventory (the high bit indicates $x$'s discourse role). 
To learn latent discourse, we adopt the similar VAE-based process as topic modeling with both the input and output as utterance $x$'s BoW ($x^{BoW}$).
First, $x^{BoW}$ is encoded into its latent discourse role $d$ with the following formula:

\begin{equation}\small
\setlength{\abovedisplayskip}{-1pt}
   \pi = gs(f_\pi(x^{BoW})), \, d=Multi(\pi)
\end{equation}
\noindent where $gs$ refers to Gumbel softmax function~\cite{Lu:17} to encode the discrete nature of latent discourse $d$ and $f_\pi$ is another neural perceptron. Afterwards, the decoding process reconstructs $x^{BoW}$ conditioned on $d$ with another fully connected layer:

\begin{equation}\small
\setlength{\abovedisplayskip}{-1pt}
    x^{BoW}=f_{\Phi^D}(d)
\end{equation}

Here similar to latent topics, we utilize $f_{\Phi^D}$'s weights to compute discourse-word distributions.

\subsection{Initiation-Response Pair Prediction}\label{ssec:model:pair-pred}

Given topic and discourse representations of a response $r$ ($z_r$ and $d_r$) and those of its candidate initiation $q$ ($z_q$ and $d_q$), we further predict how likely they form an initiation-response pair with an utterance matching process.
Here we measure the effects of topic consistency and discourse dependency to indicate initiation-response relations.

For topic consistency, we capture how similar the topics of $q$ and $r$ is with the following score:

\begin{equation}\small\label{eq:topic-score}
\setlength{\abovedisplayskip}{-1pt}
    S_{topic}(q,r)=z_r^TW_tz_q
\end{equation}
\noindent where $W_t$ is a weight matrix learned to indicate the importance of each topic factor.

Likewise, $q$ and $r$'s discourse-level matching score is denoted as $S_{discourse}$ and defined below:

\begin{equation}\small\label{eq:disc-score}
\setlength{\abovedisplayskip}{-1pt}
    S_{discourse}(q,r)=d_r^TW_dd_q
\end{equation}
\noindent where the trainable weight matrix $W_d$ is employed to capture the transition probabilities from $q$'s discourse role to $r$'s ($Pr(d_r\,|\,d_q)$).

Further, to yield the final matching score $S(q,r)$ to estimate how likely $r$ responding to $q$, we leverage $S_{topic}(q,r)$ and $S_{discourse}(q,r)$ to couple both topic and discourse effects with the weighted sum:

\begin{equation}\small\label{eq:match}
\setlength{\abovedisplayskip}{-1pt}
    S(q,r)=\gamma S_{topic}(q,r) +(1-\gamma) S_{discourse}(q,r)
\end{equation}

\noindent where $\gamma\in [0,1]$ is the parameter balancing the relative contributions of topic and discourse. 

\subsection{Learning Objectives}\label{ssec:model:objective}

Here we describe the learning objectives of the entire framework to jointly explore latent topic and discourse, and their combined effects to predict initiation-response pairs. 

\paragraph{Latent Topics and Discourse Modeling Loss.} 

We employ neural variational inference to approximate the posterior distributions over the latent topic $z$ and the latent discourse $d$. 

\textit{Encoding Topics and Discourse.}
To examine how to learn topics and discourse, the cross entropy loss is used to reflect the estimation of $z$ and $d$ from encoding process:

\begin{equation}\small
\setlength{\abovedisplayskip}{-1pt}
L_{t}=E_{q(z\,|\,c)}[\log p(c\,|\,z)]-KL(q(z\,|\,c)\,||\,p(z))
\end{equation}
\begin{equation}\small
\setlength{\abovedisplayskip}{-1pt}
L_{d}=E_{q(d\,|\,x)}[\log p(x\,|\,d)]-KL(q(d\,|\,x)\,||\,p(d))
\end{equation}
\noindent $KL$ cost term is added to avoid posterior collapse.
For space limitation, we leave out the derivation details and refer the readers to~\citet{zhao2018unsupervised}.

\textit{Reconstructing Utterances.}	
For the reconstruction loss to reflect how an utterance can be inferred from $z$ and $d$, we define the loss $L_x$ as:

\begin{equation}\small
\setlength{\abovedisplayskip}{-1pt}
L_{x}=E_{q(z\,|\,x)q(d\,|\,c)}[\log p(x\,|\,z,d)]
\end{equation}

\textit{Distinguishing Topics and Discourse.}
As discussed above, topics and discourse are modeled in different granularity (discourse in utterance only while topics in richer contexts).
To further distinguish their respective word distributions, we follow \citet{zeng2019you} to employ the mutual information to define the mutual dependency of latent topics and discourse:

\begin{equation}\small\label{eq:mi}
\setlength{\abovedisplayskip}{-1pt}
E_{q(z)q(d)}[\log \frac{p(z,d)}{p(z)p(d)}]
\end{equation}
\noindent The mutual information loss is adopted to separate the semantic space of topics and discourse:\footnote{All distributions in Eq. \ref{eq:mi} and \ref{eq:mil} are conditional distribution given utterance $x$ and its conversation context $c$ and the conditions are omitted for simplicity.}

\begin{equation}\small \label{eq:mil}
\setlength{\abovedisplayskip}{-1pt}
L_{MI}=E_{q(z)q(d)}[{KL}(p(d\,|\,z)\,||\,p(d))]
\end{equation}

\paragraph{Initiation-Response Pair Prediction Loss.}

To allow positive pairs to obtain higher scores than negative, we use hinge loss in training: 

\begin{equation}\small
\setlength{\abovedisplayskip}{-1pt}
L_{m}=\sum_{i=1}^{u} max(0,\lambda -S(q^{+},r)+S(q_i^{-},r))
\label{equa_L}
\end{equation}
\noindent where $u$ is the number of negative initiations for each response. $\lambda$ is a margin parameter and $S(q^{+},r)$ and $S(q_i^{-},r)$ are the matching scores of a response and its positive and negative initiations.

\paragraph{The Final Objective.} Finally, we combine all the effects above and define the overall objective of the entire model as:

\begin{equation}\small
\setlength{\abovedisplayskip}{-1pt}
L= L_t+ L_d+ L_x+L_m-L_{MI}
\end{equation}

In the training process, the optimization of final objective $L$ enables the end-to-end exploration of topic and discourse representation and their joint effects to signal pairwise initiation-response relations in conversation structure.

\section{Experimental Setup}

\paragraph{Data Preprocessing.} 
For \textit{CMV} dataset, the raw data was preprocessed by \citet{Tan:16}.
We first filter out tokens occurring less than $15$ maintain a vocabulary with $15,182$ tokens. 
Then, we remove utterances with word length less than $7$ or over $45$. Next, to form context for quotations and replies ($c_q$ and $c_r$),  we consider all utterances in the original post (from \textit{OH}) as $c_q$ and those in the challenger's comment as $c_r$.
Lastly, the training and test data is separated following \citet{Tan:16}, where $6,839$ pairs are used for training and and $1,098$ for test.

For \textit{CS} dataset, we don't remove words and the vocabulary size is $15,407$, with the scale similar to \textit{CMV}.
Short utterances with less than $5$ words are removed. The Chinese word segmentation and the separation of training and test set has been done by~\citet{he:2019}, with $3,701$ and $576$ instances for training and test. Here all utterances in the dialogue thread are used to form both $c_q$ and $c_r$ due to the synchronous nature of \textit{CS} conversations.

For both datasets, $10\%$ data is further sampled from the training set for validation. 

\paragraph{Model Settings.}
The hyperparameters are tuned on validation set.
For the number of topics ($K$) and discourse roles ($D$),  we set $K=50,D=5$ for \textit{CMV} dataset and $K=10,D=3$ for \textit{CS}.
Max margin weight $\lambda$ is set to $10$ (Eq. \ref{equa_L}) and $\gamma=0.5$ for balancing topic consistency and discourse dependency (Eq. \ref{eq:match}). 
In model training, we set the batch size to $32$, dropout probability to $0.5$, and the maximum epoch number to $200$ (with early stop). 
The trainable parameters are optimized via stochastic gradient descent with learning rate decay, whose initial learning rate is set to $0.1$.


\paragraph{Comparison Models.}
We first consider three non-neural baselines that rank initiations based on: 1) \underline{Position}, where earlier utterances are ranked higher for \textit{CMV} while later is higher for \textit{CS} based on the findings from Figure~\ref{fig_sta};
2) \underline{Embedding\_{Sim}} --- the cosine similarity between a response and an initiation utterance measured by the average word embeddings from Glove;
3) \underline{LDA\_Disc} --- using cross entropy to discriminate initiation's and response's topic distributions inferred by latent Dirichlet allocation (LDA)~\cite{du2017discovering}. 

We also compare with the following neural models proposed by previous work: 
1) \underline{MaLSTM}~\cite{mueller:16} designed for sentence-level semantic matching (LSTM for utterance encoding and Manhattan distance for matching);
2) \underline{CoAttention}~\cite{DBLP:conf/coling/JiWHLZH18} proposed for pairwise argument quality evaluation, where a co-attention network learns alignment representations and a BiGRU layer computes similarity for matching. 
3) \underline{RPN}~\cite{he:2019}, the state-of-the-art model for question-answer pairing in dialogues that ranks initiations by recurrent pointer networks (RPN). 

In addition, we consider the following models with a fully connected layer to score initiation-response pairs and the following encoders for utterance-level representation learning: RNN (henceforth \underline{Match\_{RNN}}), autoencoder (henceforth \underline{Match\_{AE}}), variational autoencoder (henceforth \underline{Match\_{VAE}}), and discrete variational autoencoder~\cite{zhao2018unsupervised} (henceforth \underline{Match\_{DVAE}}). 

Further, to study the relative contributions of topic consistency and discourse dependency, we compare with our two ablations, one only explores the topic effects (henceforth \underline{Topic\_Only}) and the other discourse (henceforth \underline{Discourse\_Only}).

\section{Results and Discussions}

We first discuss the main comparison results in Section \ref{ssec:results:main}, followed by an  analysis of topic and discourse effects in Section~\ref{ssec:results:effects}.
Lastly, in Section \ref{ssec:results:discussion}, we probe into our outputs and present a parameter analysis, case study, and model extension results on argument persuasiveness prediction.

\begin{table}[t!]\scriptsize
\begin{center}
\begin{tabular}{|p{1.9cm} | p{0.45cm} p{0.45cm} p{0.45cm}|  p{0.45cm}  p{0.45cm} p{0.45cm} | }
\hline
    \multirow{2}*{Models}
   &\multicolumn{3}{c|}{CMV Dataset}  &\multicolumn{3}{c|}{CS Dataset}  \\ 
   \cline{2-7}
 & Hits@1 & Hits@2 & MRR & Hits@1 & Hits@2 & MRR \\ \hline
\textbf{\underline{Non-Neural Models}} & & & && &\\
Position &24.68  &24.68   &24.68   & 49.13  & 49.13   & 49.13   \\ 
Embedding\_{Sim} & 22.77  & 45.00   & 48.66   &  17.01   &39.06   & 44.04   \\ 
LDA\_{Disc} & 24.68  & 42.99  & 47.77  & 26.39    & 49.65   & 52.40  \\ 
\hline\hline
\textbf{\underline{Neural Models}} & & &  && &\\
MaLSTM & 29.87  & 42.99    & 50.91  & 43.58    &72.40   & 65.80  \\ 
CoAttention & 47.72  & 68.31   & 67.26   & 51.56    &79.17    & 71.77   \\ 
RPN & 46.45  & 67.21    & 66.22   & 52.95   & 80.21  & 72.69   \\ 
Match\_{RNN} &49.45   & 71.58  & 68.79  &50.00     &80.38   &71.13    \\ 
Match\_{AE} &51.82   &$\textrm{74.77}$   &70.50    &52.78    &$\textrm{82.12}$   & 72.88   \\ 
Match\_{VAE}&53.19   &$\textrm{73.95}$ &$\textrm{71.11}$  & 52.60    & 81.42  &72.70    \\ 
Match\_{DVAE} &47.45  &69.95   & 67.34   &53.82     &$\textrm{82.81}$  &73.65   \\ 
\hline\hline
\textbf{\underline{Ablations}} & & &  && &\\
Topic\_{Only} &58.20  &76.14 &73.78   &42.53     &69.10    &64.11   \\  
Discourse\_{Only} &41.44   &63.02    &62.20    &48.96   &76.04   &69.76    \\ 
\hline
\hline
\textbf{Our model} &\textbf{59.74}  &\textbf{76.23}  &\textbf{74.41}   & \textbf{64.93}   &\textbf{84.20}    &\textbf{79.23}   \\  \hline
\end{tabular}
\end{center}
   \vspace{-1em}
\caption{Comparison results on two datasets and our model achieves the best results under all settings. Our model significantly outperforms the comparison model (Wilcoxon signed rank test, \emph{p}$<$0.05). }
\vspace{-1.5em}
\label{tab:tb2}
\end{table}

\subsection{Main Comparison Results}\label{ssec:results:main}

The overall results are shown in Table~\ref{tab:tb2}. Two widely-used information retrieval metrics \textit{Hits@N}, $N=1,2$ and Mean Reciprocal Rank (\textit{MRR}) are used for evaluation metrics. Several interesting observations can be drawn.

\noindent~$\bullet$  \textit{All models yield generally better performance on CS than CMV.}
It shows that initiation-response links are more difficult to be identified on dialogues in argumentative than everyday styles. 

\noindent~$\bullet$ \textit{Neural networks perform better than non-neural baselines.} 
Shallow features from position, word embeddings, and LDA-based latent topics can't obtain good performance. 
Neural models explore deeper semantic features and provide better results. 

\noindent~$\bullet$ \textit{Autoencoders learn useful representations.} It is observed that models based on autoencoders perform generally better than other models. This shows that autoencoders are effective in encoding utterances compared with other alternatives.
 
\noindent $\bullet$ 
\textit{Topics contribute more on \textit{CMV} while discourse is more useful in \textit{CS}.} Topic\_Only performs much better than Discourse\_Only on \textit{CMV}, while the opposite is observed on \textit{CS}.
It is probably because of the richer context in \textit{CMV} to learn latent topics (with more words per conversation as shown in Table \ref{tab:tb1}), while the synchronous \textit{CS} dialogues exhibits richer discourse word patterns from back and forth interactions between participants and hence allow better discourse modeling. 

\noindent $\bullet$ \textit{Our model significantly outperforms all comparisons.} This shows that the joint effects of topics and discourse can usefully indicate the relations of initiations and responses in conversation context. 
\begin{figure}[t]
\subfigure{
\includegraphics[width=0.45\columnwidth] {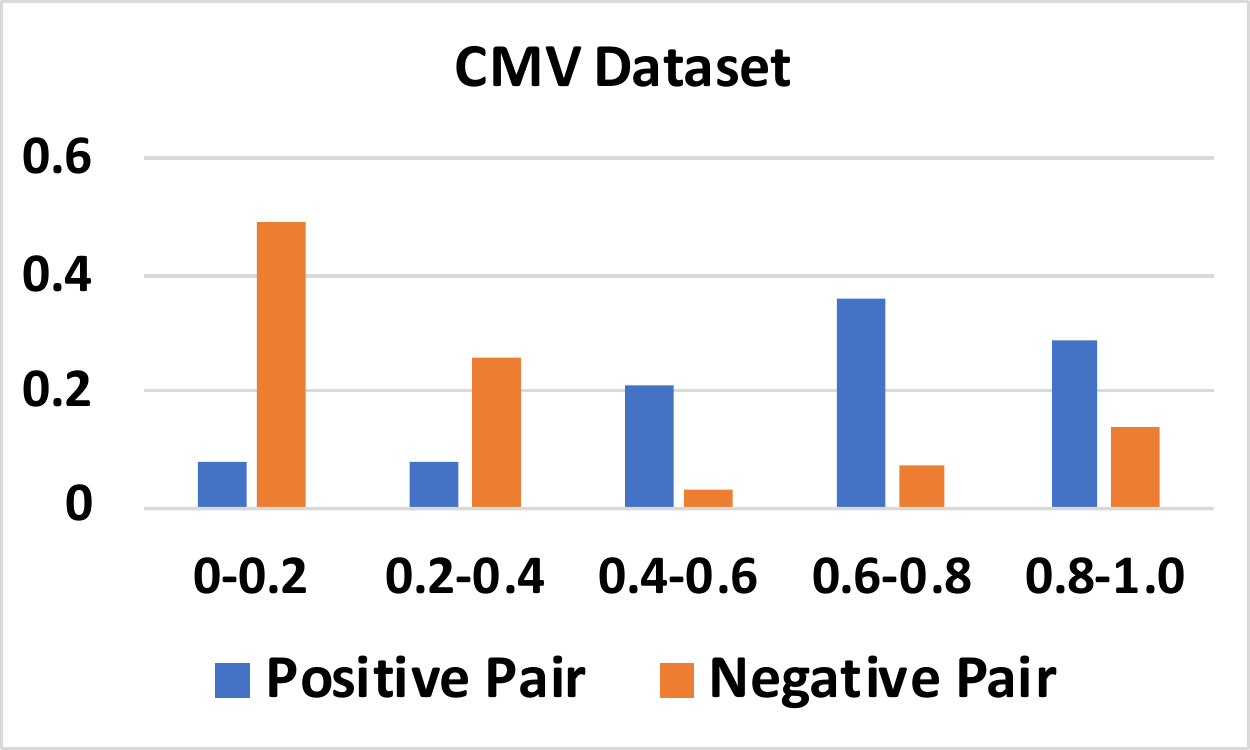}
}
\subfigure{
\includegraphics[width=0.45\columnwidth] {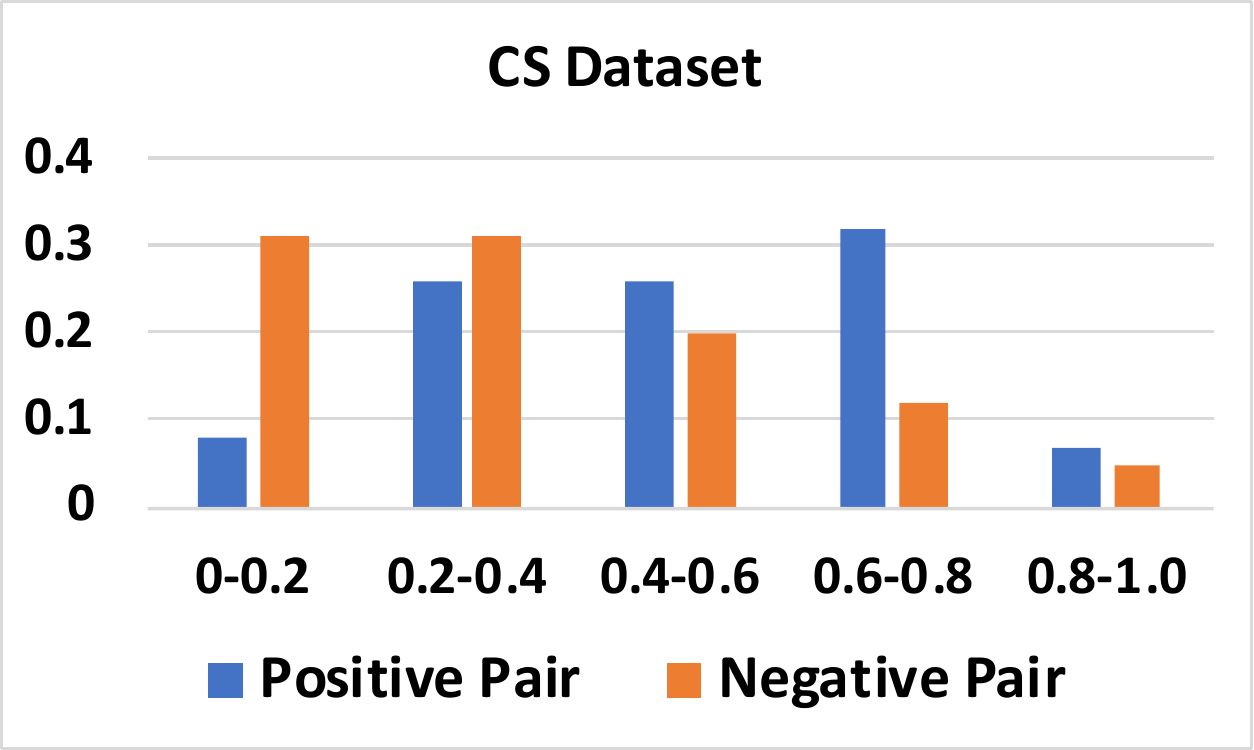}
}
\vspace{-1em}
\caption{\label{fig_topic}The distribution of topic similarity in the CMV dataset (a) and CS (b). X-axis shows cosine similarity intervals and y-axis indicates proportions. For each interval, positive pair results are displayed on the left (in blue) and negative on the right (in orange). }
\vspace{-1.5em}
\end{figure}

\subsection{Effects of Topics and Discourse}\label{ssec:results:effects}


\paragraph{Topic Effects.} 
We first analyze the effects of topic consistency and compute the cosine similarity of the latent topics we learn for responses ($z_r$) and candidate initiations ($z_q$). 
The distributions over positive and negative pairs are shown in Figure \ref{fig_topic}.
For both datasets, our model generally assigns higher topic similarity for positive pairs than negative, probably because responses tend to follow the concern of initiations and are hence likely to contain similar topic words.
We also observe a proportion drop in very similar positive pairs ($sim>0.8$), indicating that most responses do not echo what were said in initiations, though their topics might be similar. 
Nevertheless, negative pairs exhibit different distributions compared with the positive ones.
Our model is able to capture such features in topic consistency modeling (Eq. \ref{eq:topic-score}), which might help in distinguishing positive and negative initiations for a response.
\begin{figure}[t]
\subfigure{
\includegraphics[width=0.47\columnwidth] {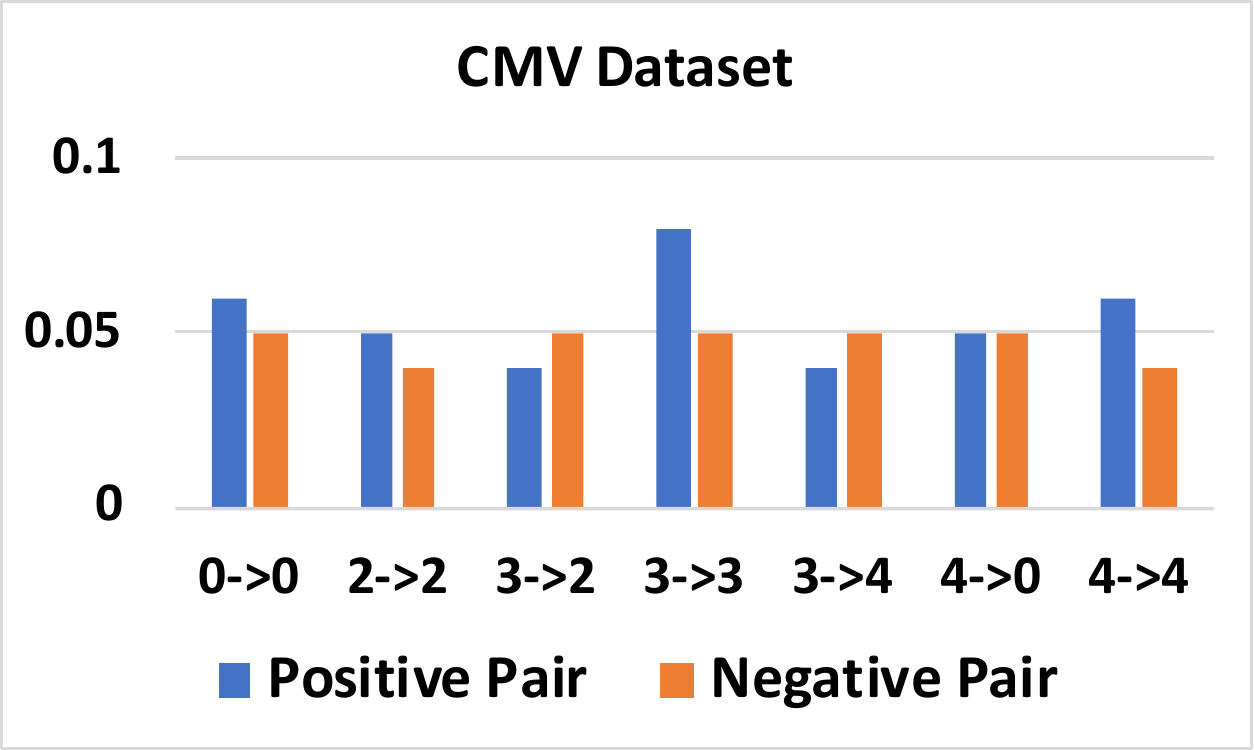}
}
\subfigure{
\includegraphics[width=0.47\columnwidth] {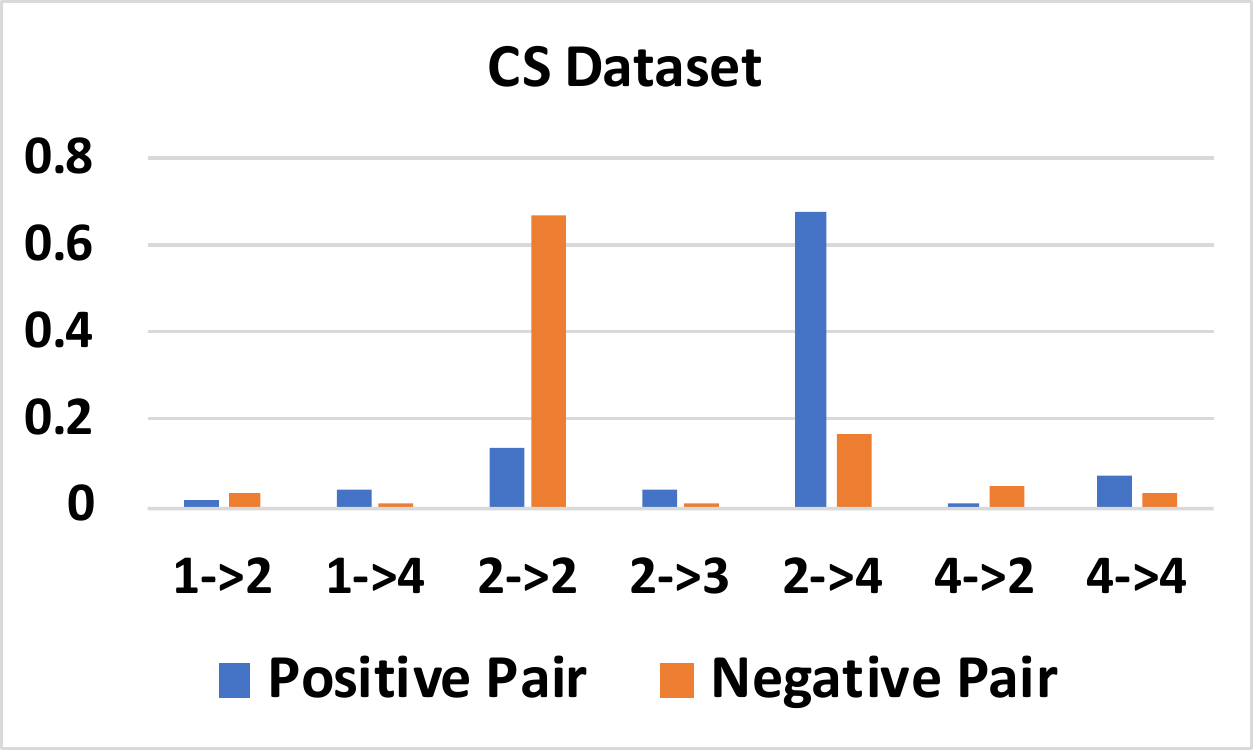}
}
\caption{\label{fig_discourse}The transition distributions of discourse roles from initiations to responses, \textit{CMV} in (a) and \textit{CS} in (b). 
Only the top $5$ transitions observed in positive (on the left in blue) and negative pairs (on the right in orange) are displayed. X-axis: initiation-response discourse roles ($d_q\rightarrow d_r$); Y-axis: proportions.}
\vspace{-1em}
\end{figure}

\paragraph{Discourse Effects.} We discuss how discourse dependency affects the prediction of initiation-response pairs. 
The transition distributions of discourse roles from initiations to responses ($d_q\rightarrow d_r$) are shown in Figure \ref{fig_discourse}.
As can be seen, the discourse transition distributions in \textit{CS} dataset are diverse for positive and negative pairs. It may help explain why discourse can better signal initiation-response pairs on \textit{CS} compared with \textit{CMV} (observed from Discourse\_Only's performance in Table \ref{tab:tb2}).
For \textit{CMV}, there are slightly different distributions for positive and negative pairs.
For this reason, topic factors may contribute more than discourse (seen via comparing Discourse\_Only and Topic\_Only on \textit{CMV}).
This also indicates that discourse modeling for argumentative dialogues is challenging, which may require the learning of more complex features other than word statistics and is beyond the capacity of our model.

\subsection{Further Discussions}\label{ssec:results:discussion}

\paragraph{Parameter Analysis.} Here we investigate the impact of two hyper-parameters on our model, namely, the number of topics ($K$) and discourse ($D$).

\begin{figure}[t!]
\subfigure{
\includegraphics[width=0.48\columnwidth] {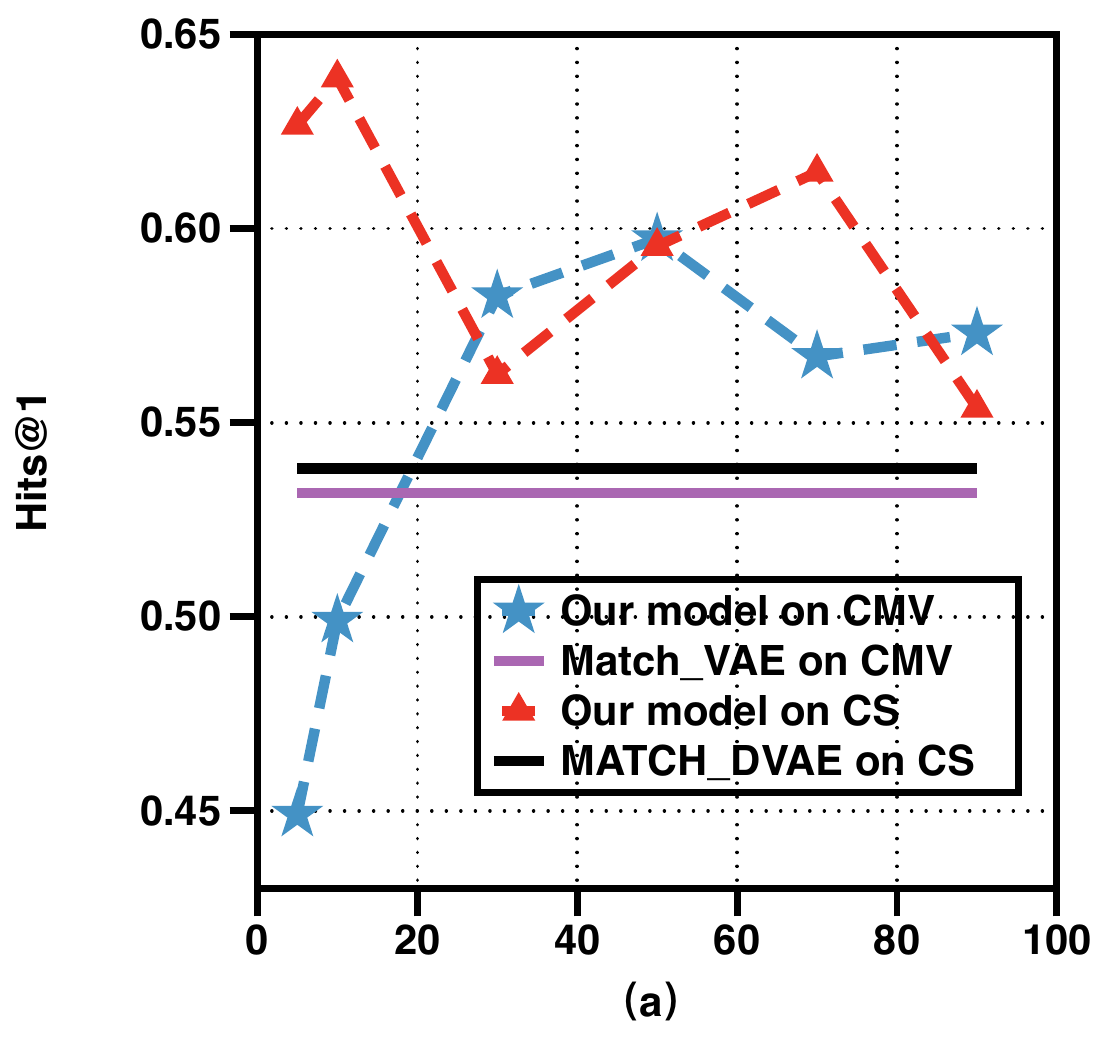}
}\hspace{-3mm}
\subfigure{
\includegraphics[width=0.48\columnwidth] {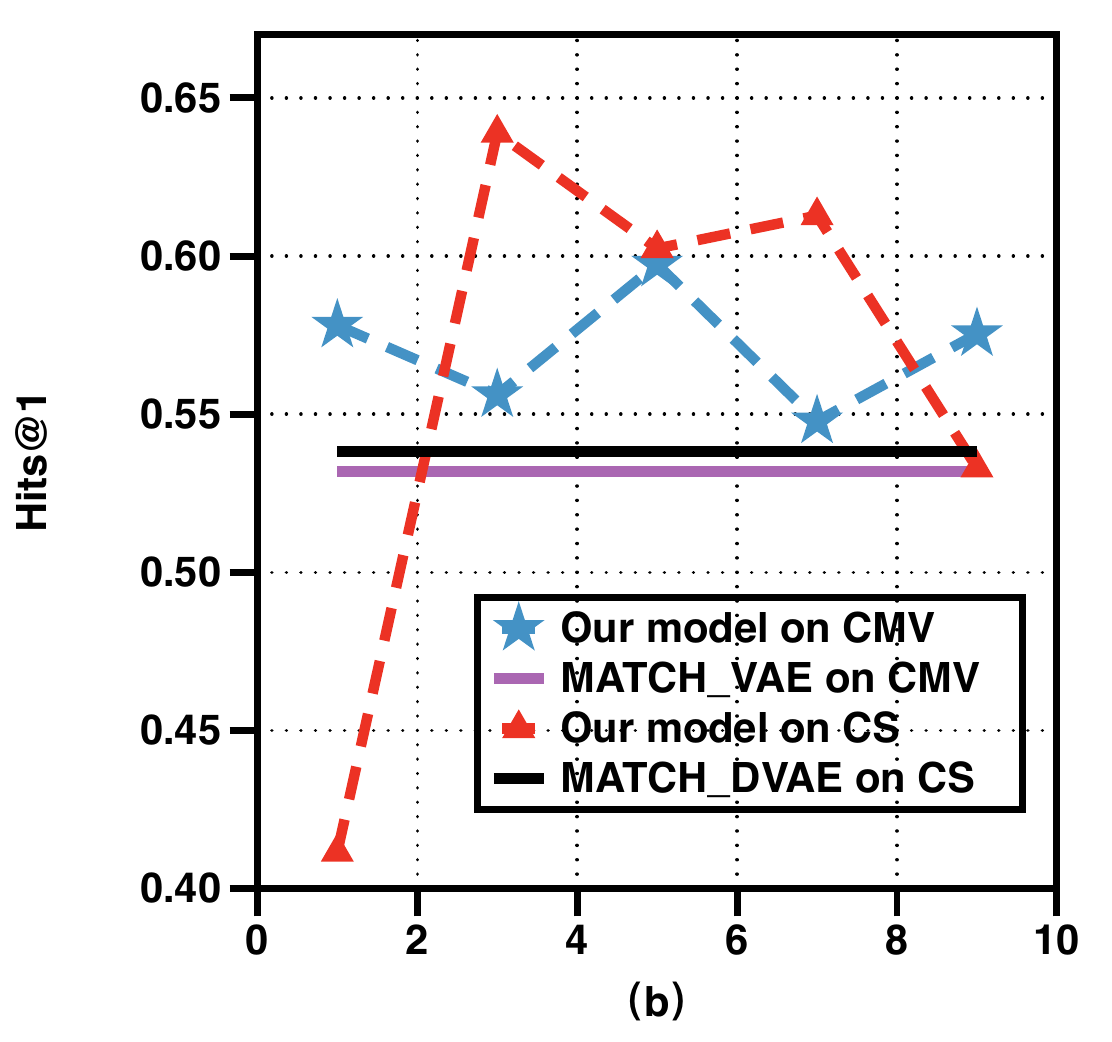}
}
\vspace{-1em}
\caption{\label{fig1}Hits@1 over varying number of topics and discourse 
X-axis: topic number ($K$ in (a)) and discourse number ($D$ in (b)). 
Y-axis: Hits@1 score.
Blue and red curves: our model on \textit{CMV} and \textit{CS}.
Purple and black lines: Match\_DVAE on \textit{CMV} and \textit{CS}. }
\vspace{-1em}
\end{figure}

\textit{Varying Topic Number.} Figure \ref{fig1} (a) shows how Hits@1 scores change over varying number of topics ($K$). 
For comparison, we also display  Match\_DVAE's results, the best comparison model in Table \ref{tab:tb2}.
For relatively large $K$, our model performs consistently better than Match\_DVAE.
We also find that our trend on both datasets are not monotonic, where the best performance is attained at $K=50$ for \textit{CMV} and $K=10$ for \textit{CS}.
This implies that the topics in customer service dialogues are limited while participants may discuss wide range of topics in social media debates.

\textit{Varying Discourse Number.} The results for varying discourse number ($D$) are displayed in Figure \ref{fig1} (b). Similar to $K$, our model exhibits better results than Match\_DVAE for $D>1$.
It's also observed that \textit{CS} is more sensitive to $D$ compared with \textit{CMV}, indicating that discourse factors largely affect the initiation-response prediction results on $CS$.

\begin{figure}[t!]
\centering
   \includegraphics[width=0.9 \columnwidth ]{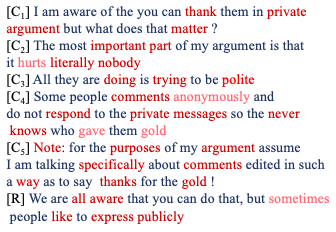}
   \vspace{-1em}
    \caption{\label{fig21}Visualization of the topic and discourse word assignment for the \textit{CMV} conversation snippet in Figure \ref{fig11}. The blue words are prone to indicate discourse ($p(w\,|\,d)>p(w\,|\,z)$) while red topic. Darker colors indicate higher confidence.}
\vspace{-1em}
\end{figure}

\paragraph{Case Study.} To examine what we learn to represent topics and discourse, we take the \textit{CMV} conversation snippet in Figure~\ref{fig11} as an example to analyze the topic and discourse words assigned by our model.  
Recall that $R$ answers $C_1$'s question suggested by the shared pronoun ``that'' and the similar topics they concern.
Figure~\ref{fig21} shows the visualization results and displays topic words in red and discourse in blue.
It is observed that our model is able to separate topic words (e.g., ``thank'', ``private'', and ``public) from discourse (e.g., ``that'', ``what'', and ``?''), which may result in coherent topic and discourse distributions and indicative  representations to signal initiations-response relations. 
Interestingly, discourse words are mostly stop words and punctuation. 
Their meaningful clusters exhibiting different statistic patterns might usefully indicate varying discourse behaviors in conversations, which is consistent with the findings from previous studies~\cite{DBLP:journals/coling/LiSWW18,zeng2019you}.

\paragraph{Downstream Task.} 
In Introduction, we mentioned that the detection of initiation-response pairs may contribute to a better understanding of conversation structure and hence benefit downstream applications. 
We take the prediction of argument persuasiveness as an example to discuss whether the representations learned by our model can advance the state-of-the-art performance on this task. 
Table \ref{tab:tb15} shows the performance of the non-neural baseline \cite{Tan:16}, the state-of-the-art model \cite{DBLP:conf/coling/JiWHLZH18}, and \citet{DBLP:conf/coling/JiWHLZH18} incorporating the topic and discourse representations we learn ($z$ and $d$).
The dataset is collected from \textit{CMV} and argument quality is labeled by $\Delta$ (given by \textit{OH} to indicate the successful persuasion).
It is seen that the latent topics and discourse learned to signal initiation-response relations can help to predict argument quality, suggesting that the persuasiveness of arguments are closely related to the structure of who respond to whom in argumentation processes.

\section{Related Work}

Our work is in the line with prior efforts to detect initiation-response pairs. \citet{wang:2010} explore how topic features discovered via latent semantic analysis work in this task, ignoring the effects of discourse roles. However, our study shows that both topics and discourse are helpful to identify who respond to whom in conversation structure.
Other work~\cite{jamison2014adjacency,du2017discovering,chen2017learning} focus on the design of hand-crafted features (e.g., text similarity, pre-detected dialogue acts, etc.).
Compared with them, our model enables learning of deep semantic features for topics consistency and discourse dependency, which can be captured without the labor-intensive feature engineering process. 
Recently, there exists an attention over how neural framework perform to identify replying relations in conversation discourse \cite{guo2018answering,he:2019}.
However, they ignore the effects of latent topics and discourse to structure a conversation, which are studied here and shown useful to indicate initiation-response relations in experiments.

We are also inspired by the previous approaches to discover latent topics and discourse in conversations contexts. 
Many of them employ probabilistic graphical models in LDA-fashion to explore word statistics~\cite{ritter2010unsupervised,DBLP:journals/coling/LiSWW18,zeng2018microblog}. 
These models use expertise involvements to customize inference algorithms, whereas our architecture allows end-to-end training of topic and discourse model together with other components. We take the advantage of the recent progress to explore conversation representations via variational autoencoders~\cite{miao:2017,zhao2018unsupervised,zeng2019you}, allowing to capture topic and discourse factors in an unsupervised manner.
However, their effects to signal user interactions in conversation structure have never been studied before, which is a gap our work fills in.
\begin{table}[t]\footnotesize
\begin{center}
\begin{tabular}{ |c | c |}
\hline Models & Pairwise accuracy   \\ \hline
~\citet{Tan:16} (\textit{baseline}) & 65.70    \\ 
~\citet{DBLP:conf/coling/JiWHLZH18} (\textit{SOTA}) & 70.45\\
~\citet{DBLP:conf/coling/JiWHLZH18}+Our model & 74.12 \\ \hline
\end{tabular}
\end{center}
\vspace{-1em}
\caption{\label{tab:tb15}The pairwise accuracy to predict argument persuasiveness. The results in the first two rows were reported in their original paper. Our representations help advance the state of the art (SOTA).}
\vspace{-1.5em}
\end{table}
\section{Conclusion}

This work explores the effects of latent topics and discourse roles to signal initiation-response relations.
We first employ a VAE-based neural model to capture topic and discourse representations in an unsupervised manner.
Then, topic consistency and discourse dependency are further exploited to predict how likely an utterance responds to an initiation.
Extensive experiments on two datasets 
show that our model significantly outperform the previous state-of-the-art models.
Further analyses show that both topics and discourse are useful to signal who respond to whom in conversation structure. 

\bibliography{anthology,custom}
\bibliographystyle{acl_natbib}

\end{document}